\documentclass{article}

\pdfoutput=1

\newcommand{\etal}{\textit{et al}.}

\newcommand{\Tref}[1]{Table~\ref{#1}}

\newcommand{\fref}[1]{Fig.~\ref{#1}}
\newcommand{\Fref}[1]{Figure~\ref{#1}}

\PassOptionsToPackage{numbers, compress}{natbib}
\usepackage[final]{nips_2017}

\usepackage[utf8]{inputenc} % allow utf-8 input
\usepackage[T1]{fontenc}    % use 8-bit T1 fonts
\usepackage{hyperref}       % hyperlinks
\usepackage{url}            % simple URL typesetting
\usepackage{booktabs}       % professional-quality tables
\usepackage{amsfonts}       % blackboard math symbols
\usepackage{nicefrac}       % compact symbols for 1/2, etc.
\usepackage{microtype}      % microtypography
\usepackage{graphicx}
\usepackage{subfigure}
\usepackage{amsmath}

\title{Deep Semantics-Aware Photo Adjustment}

\author{
  Seonghyeon Nam \\
  Department of Computer Science \\
  Yonsei University \\
  \texttt{shnnam@yonsei.ac.kr} \\
  \And
  Seon Joo Kim \\
  Department of Computer Science \\
  Yonsei University \\
  \texttt{seonjookim@yonsei.ac.kr} \\
}

\begin{document}

\maketitle

\begin{abstract}
  Automatic photo adjustment is to mimic the photo retouching style of professional photographers and automatically adjust photos to the learned style. There have been many attempts to model the tone and the color adjustment globally with low-level color statistics. Also, spatially varying photo adjustment methods have been studied by exploiting high-level features and semantic label maps. Those methods are semantics-aware since the color mapping is dependent on the high-level semantic context. However, their performance is limited to the pre-computed hand-crafted features and it is hard to reflect user's preference to the adjustment. In this paper, we propose a deep neural network that models the semantics-aware photo adjustment. The proposed network exploits bilinear models that are the multiplicative interaction of the color and the contexual features. As the contextual features we propose the semantic adjustment map, which discovers the inherent photo retouching presets that are applied according to the scene context. The proposed method is trained using a robust loss with a scene parsing task. The experimental results show that the proposed method outperforms the existing method both quantitatively and qualitatively. The proposed method also provides users a way to retouch the photo by their own likings by giving customized adjustment maps.
\end{abstract}

\section{Introduction}
With the growing number of digital cameras especially with smartphones, photo retouching softwares have become popular among amateur photographers. As the captured photos are usually flat, many people want to adjust the tone and the color of the photos, to make the pictures to look visually more impressive and even stylized. However, the photo retouching is a hard task for the amateur users without the expertise in the photo editing. Additionally, retouching a large photo collection requires extensive human labor.

For this reason, many techniques for automatic photo adjustment have been widely studied. The automatic photo adjustment automatically enhances photos' tone and color to be visually more pleasing without human actions. 
In the automatic photo retouching, the output styles mimic the photo styles of professional photographers.
Several methods have been proposed to adjust the contrast/brightness and the color/saturation of photos~\cite{bychkovsky2011learning,kapoor2014collaborative} based on low-level color histogram, the brightness, and the contrast of images. 
However, those methods adjust photos globally by applying the same color mapping to all pixels in an image. 
Note that most photographers prefer locally varying adjustments in their work.

Some works have focused on spatially varying photo adjustment that exploits high-level scene contexts based on the object features and the saliency~\cite{hwang2012context,Yan16}.
In~\cite{Yan16}, the authors use a feed-forward neural network to learn the semantics-aware photo adjustment styles of professional photographers.
In the semantics-aware photo adjustment, the tone and the color mapping are dependent on the scene context, which is a local regions of a given image.
The authors proposed multi-scale pooling features of the semantic label map to model the context dependency. 
However, the work uses hand-designed features, and it is unclear whether their hand-designed features based on inaccurate semantic label map are optmial. In addition, the learned representation of the method is not separated, and therefore  users cannot control the adjustment by their own preference.
%Also, their neural network is a kind of blackbox and hard to understand what is going on inside the neural network.

In this paper, we propose a deep neural network (DNN) that learns the representation of the semantics-aware photo adjustment in an end-to-end manner. 
While we make use of the dataset from~\cite{Yan16}, we approach the problem in a different way.
First, the proposed network is trained in an end-to-end manner so that it fits better to the data.
Our network is a bilinear model where the color and the contextual information is interacted in a multiplicative way.
We exploit multi-scale convolutional neural network (CNN) features to characterize pixel-wise contextual features.
Unlike~\cite{Yan16}, the contextual features are learned within the network in an end-to-end manner.
To efficiently train the network, we make use of a robust loss function and the multi-task learning with a scene parsing task.
Second, as another type of contextual features, we introduce a semantic adjustment map.
The semantic adjustment map is a binary segmentation map that discovers the photo retouching presets which vary according to the semantic contexts. The network automatically disentangles different types of presets from the original in an unsupervised manner and adjust images accordingly.
By doing so, we can understand better the photo retouching styles and use the discovered presets to adjust the photos for each user's preference.
Note that our photo adjustment framework is different from the image style transfer~\cite{Gatys_2016_CVPR} that stylizes photos to look like artworks. 
Instead of focusing on the global modification of shapes and textures, we focus on the tone and the color manipulation of images.

\section{Related works}
There has been a number of studies for the automatic photo adjustment.
Several methods focus on the global tonal adjustment~\cite{bychkovsky2011learning}, the color enhancement~\cite{yan2014learning}, and the personalized enhancement~\cite{kapoor2014collaborative}.
Those methods are global adjustment approaches based on hand-crafted low-level features such as the color histogram, the scene brightness, and the highlight clipping.
In~\cite{kapoor2014collaborative}, Kapoor~\etal~proposed a method that discovers the clusters of users that have similar preferences of image enhancement for the personalized adjustment.
While the concept of our method may be similar to those methods, the main difference is that we aim to discover the retouching presets that vary according to the local semantics.

Hwang~\etal~\cite{hwang2012context} presented a locally varying photo enhancement method that is based on both low- and high-level contexts. Their method finds an appropriate color mapping from external images using pixel-wise contextual features. The work of Yan~\etal~\cite{Yan16} is closely related to our work. The authors combine multiple hand-crafted features including a multi-scale pooling of a scene parsing map for semantics-aware color regression. While the multi-scale pooling features were effective in modelling the semantics-aware photo adjustment, the performance is limited to the quality of the scene parsing map since the features are not trained in an end-to-end manner.

Our method is also related to various deep learning based semantics-aware image processing methods. Tsai~\etal~\cite{Tsai_SIGGRAPH_2016} used a scene parsing deep network to localize a sky region and transfer a different style of sky from external images. In~\cite{Tsai_CoRR_2017}, the authors propose a DNN for image harmonization, which is an encoder-to-decoder network to exploit high-level contextual features. The DNN is jointly trained with a scene parsing task to improve the training. In contrast to~\cite{Tsai_CoRR_2017}, our method does not rely on the segmentation mask and rather finds the inherent segmentation masks from the data. Deep learning based colorization methods~\cite{Zhang16,Larsson16} are also related to our work in that the methods make use of rich contextual features of CNNs to estimate the color of a pixel according to the scene context. Unlike those methods, we do not reconstruct missing color channels, and the color mapping of pixels is consistent in a semantic region.

\begin{figure}[h]
  \centering
  \includegraphics[width=0.85\linewidth]{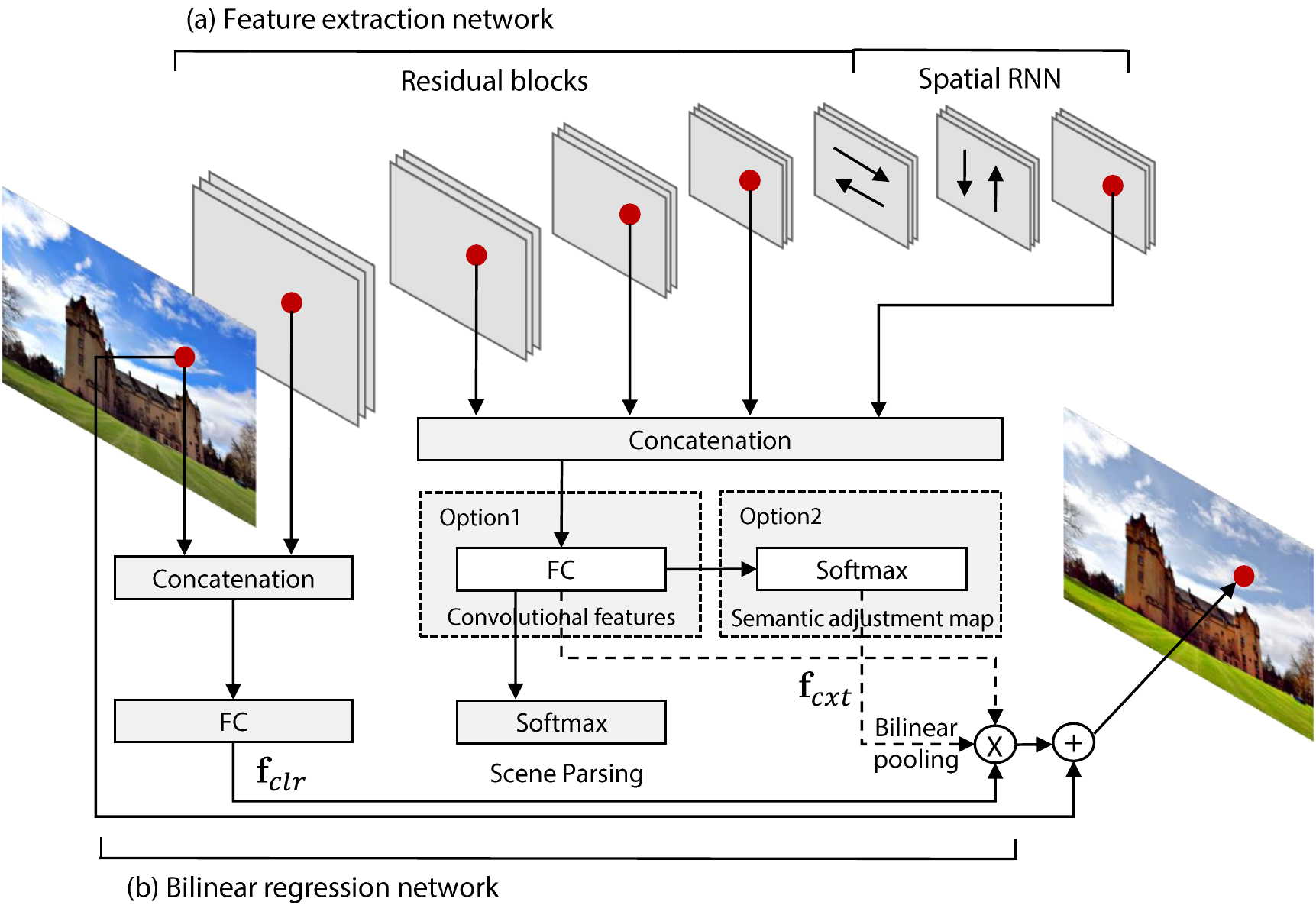}
  \caption{The overview of the propose neural network.}
  \label{fig:overview}
\end{figure}

\section{Method}
\label{sec:method}
\subsection{Overview}
We define the semantics-aware photo adjustment problem as a regression problem. We want to find a regression model of the color mapping from the input color $\mathbf{x}$ to the output color $\mathbf{y}$ according to the semantic context that the input pixel belongs to. To this end, we propose a deep neural network that effectively learns the context dependent color mapping.

\Fref{fig:overview} shows the overview of the proposed deep network. 
Our network is divided into two parts: a feature extraction network and a bilinear regression network.
The feature extraction network is based on the ResNet-50~\cite{he2016deep} as shown in \fref{fig:overview} (a). The contextual features of the ResNet-50 are effective for modelling the semantics-aware color mapping, since we can exploit low to high level pixel-wise features that are pretrained on a large dataset.
However, those convolutional features only describe the local context.
For the better context modelling, the global context and the relative compositional context between scene objects would be useful.
Therefore, we add a spatial RNN to extract those global and relative contexts.
We adopt the ReNet~\cite{visin2015renet} that consists of 4 directional spatial RNN layers, followed by an additional 1$\times$1 convolution.
To avoid the overfitting, we use GRU~\cite{cho2014properties} as a spatial RNN cell with batch normalization~\cite{ioffe15}.

The bilinear regression network shown in \Fref{fig:overview} (b)  estimates the output color given both the input color features and the contexual features. In the following, we describe the bilinear regression network in detail.

\subsection{Bilinear model}
Bilinear models are the multiplicative interaction of all elements between two vectors~\cite{tenenbaum1996separating,Kim2016c,lin2015bilinear}.
Formally, a bilinear model is defined as
\begin{equation}
f_i = \mathbf{a}^T\mathbf{W}_i\mathbf{b} = \sum_{jk} a_j b_k W_{ijk},
\end{equation}
where $\mathbf{a}$, $\mathbf{b}$ are feature vectors, and $\mathbf{W}_i$ is the interaction between two vectors.

In the semantics-aware photo adjustment, it is natural to think that the color mapping is determined by two factors; one is the color of a pixel and the other is the scene context that the pixel belongs to. Therefore, we use the bilinear model to represent the interaction between both factors. Since $\mathbf{W}\in\mathbb{R}^{C \times N \times M}$ is usually high-dimensional, we follow the low-rank bilinear pooling method of Kim~\etal~\cite{Kim2016c} to reduce the parameters. Based on the method, the output color $\mathbf{\hat{y}}$ is represented as
\begin{equation}
\mathbf{\hat{y}} = \sigma(\mathbf{P}^T(\sigma(\mathbf{U}^T\mathbf{f}_{clr} + \mathbf{b})\circ\sigma(\mathbf{V}^T\mathbf{f}_{cxt} + \mathbf{c}))+\mathbf{d}),
\end{equation}
where $\mathbf{f}_{clr}\in\mathbb{R}^{N}$ is color features, $\mathbf{f}_{cxt}\in\mathbb{R}^{M}$ is context features, $\mathbf{P}\in\mathbb{R}^{d \times c}$, $\mathbf{U}\in\mathbb{R}^{N \times d}$, $\mathbf{V}\in\mathbb{R}^{M \times d}$ are the decomposition of $\mathbf{W}$, and $\mathbf{b}\in\mathbb{R}^{d}$, $\mathbf{c}\in\mathbb{R}^{d}$, $\mathbf{d}\in\mathbb{R}^{c}$ are addtional biases. $\circ$ is an element-wise multiplication and we use $\tanh$ as a nonlinear function $\sigma$.
Note that $\mathbf{\hat{y}}$ is actually a residual since we add a skip connection between the input $\mathbf{x}$ and $\mathbf{\hat{y}}$: $\mathbf{y} = \mathbf{\hat{y}} + \mathbf{x}$.

The method of Yan~\etal~\cite{Yan16} exploits an asymmetric form of bilinear model~\cite{tenenbaum1996separating} by estimating affine transformaion matrices to map quadratic color features to output colors. On the other hand, our method is more flexible and efficient in that our bilinear model learns the nonlinear interaction of two features as well as both feature representations. For both cases, it is clear that merging two features in a multiplicative manner is beneficial for the semantics-aware photo adjustment.

\subsubsection{Color features}
We use the CIELab color space for both the input and output images. We can use 3-channel Lab color as the color features. However, it generates color variations in smooth regions since each color is processed independently. To alleviate this issue, we add the local neighborhood information by concatenating the Lab color and the $L_2$ normalized first-layer convolutional feature maps of ResNet-50.

\subsubsection{Contextual features}
\paragraph{Convolutional features}
We first take advantage of the multi-scale convolutional features.
To generate pixel-wise features from the multi-scale feature maps, we adopt the sparse hypercolumn training method~\cite{BansalChen16,Larsson16}, which requires much less parameters than the deconvolutional approaches~\cite{long2015fully,noh2015learning}.
In the training time, we generate many training signals by randomly sampling sparse pixels from the image for the backpropagation. When we are given a small data, we can exploit both low to high level features efficiently with this approach.

We use the first 3 residual blocks for the hypercolumn, which have 256, 512, and 1024 channels, respectively. As mentioned, we additionally use spatial RNN features that have 1024 channels. We normalize each feature map by its $L_2$ norm, concatenate them, and squeeze the feature dimension to 512 by using 1$\times$1 convolution as shown in the option 1 of \fref{fig:overview} (b).

\paragraph{Semantic adjustment map}
As the convolutional features are unconstrained and smooth, they can represent rich scene contexts. However, two real-valued bilinear features are highly correlated, and it is difficult to understand which factor contributes to a specific style of color mapping. It would be better if we can separate those factors not only to interpret the retouching styles according to the scene contexts, but to make use of those styles for our own taste.

To this end, we generate K-channel binary maps, of which each channel is a binary segmentation map that one of the retouching presets is applied to. For each pixel, an one-hot vector $\mathbf{f}_{cxt}$ is a categorical random variable, which is defined as
\begin{equation}
\mathbf{f}_{cxt} = \mathbf{m}\sim Cat(\{p(m_1=1|\mathbf{x}), p(m_2=1|\mathbf{x}), ..., p(m_K=1|\mathbf{x})\}),
\end{equation}
where $\mathbf{m}$ is a one-hot vector sampled from a categorical probability density function $p(m_k=1|\mathbf{x})$. $p(m_k=1|\mathbf{x})$ is a probability of retouching a pixel $\mathbf{x}$ using the k-th retouching preset.
Similar to~\cite{xu2015show}, we reformulate our regression loss $\log p(\mathbf{y}|\mathbf{x})$ using a variational lowerbound technique, which is described as
\begin{equation}
\begin{split}
L_{reg} &= \log(p(\mathbf{y}|\mathbf{x})) \\
&= \log(\sum_{k=1}^K p(m_k=1|\mathbf{x})p(\mathbf{y}|m_k,\mathbf{x})) \\
&\ge \sum_{k=1}^K p(m_k=1|\mathbf{x}) \log(p(\mathbf{y}|m_k,\mathbf{x})) \\
&= \mathbb{E}_{p(m_k=1|\mathbf{x})}[\log(p(\mathbf{y}|m_k,\mathbf{x}))].
\end{split}
\end{equation}
In our task, K is typically small enough to compute the exact expectation if we assume that the pixels are independent to each other. In practice, however, it is likely that the problem converges to a local minimum that all retouching styles are classified to one or two classes. It is because the number of traininig examples for each retouching style is imbalanced. In other words, the optmization is dominanted by a few large classes such as the sky and the ground. In~\cite{eigen2015predicting}, the authors use a class reweighting trick for class-balanced classification. Similarly, we multiply different weights to each K loss term to alleviate the issue.
In contrast to~\cite{eigen2015predicting}, we multiply small weights to the loss term of low-frequency classes so that small classes are easily discovered in spite of relatively small training signals.
The weight is defined as
\begin{equation}
\mathbf{w}_t = \alpha \times \mathbf{a}_t + (1 - \alpha),
\end{equation}
where $\alpha$ controls the contribution of the weight $\mathbf{a}$ to the loss. 
$\mathbf{a}_t$ is the moving average of normalized soft frequences of K classes that is computed from $t$ training batches defined as
\begin{equation}
a^k_t = 0.9 \times a^k_{t-1} + 0.1 \times \frac{1}{P} \sum_i p^i_t(m_k=1|\mathbf{x}),
\end{equation}
where $\frac{1}{P} \sum_i p^i_t(m_k=1|\mathbf{x})$ is the average of $p_t(m_k=1|\mathbf{x})$ for all pixels in a $t$-th batch. Our final regression loss is formulated as
\begin{equation}
L_{reg} = \mathbb{E}_{p(m_k=1|\mathbf{x})}[\mathbf{w}_t \log(p(\mathbf{y}|m_k,\mathbf{x}))].
\end{equation}

\subsection{Huber loss}
To generate the ground truth of adjusted photos, photographers use a segmentation tool to localize a region of a specific object to retouch.
Although they thoroughly follow the procedure, some outliers may exist around object boundaries due to the incorrect segmentation. Also, the adjustment style of a photographer may not be consistent from an image to another image. Therefore, the optimization of our deep network should be robust to such outliers.

As a training objective, $L_2$ loss is widely used in various color regression tasks~\cite{Yan16,Larsson16}. However, DNNs easily overfit to outliers since the gradient of $L_2$ loss is large for those outlier samples and the optimization is dominanted by them. As an alternative to $L_2$, Huber loss~\cite{huber1964robust} is more robust to outliers, which is defined as
\begin{equation}
L_{huber}(e) = \left\{
\begin{array}{ll}
\frac{1}{2} e^2 & \mbox{for}~|e| \le \delta,\\
\delta (|e| - \frac{1}{2}\delta) & \mbox{otherwise},
\end{array}
\right.
\end{equation}
where $e$ is error and $\delta$ is the changepoint between the two loss functions. The loss is quadratic for a small error $|e| \le \delta$, and linear for a large error $|e| > \delta$. As the gradient of the linear function is always $\delta$, the contribution of outliers in the optimization is reduced.

\subsection{Multi-task learning}
Unfortunately, getting a large labeled dataset for the photo adjustment is not easy, since photo editing requires tremendous human labor.
When the proposed network is trained on such a small dataset, it is highly likely to overfit to a few specific scene contexts.
Since pixel-wise semantic information is the key to our semantics-aware photo adjustment, the overfitting is very severe and results in inconsistent color mappings.
To mitigate this problem, we simultaneously train a scene parsing task with our task as a regularization, thereby our deep network can be generalized to any scene contexts.

To train the scene parsing task, we use the SceneParse150 dataset~\cite{zhou2017scene}, which consists of 150 scmantic categories.
As depicted in~\fref{fig:overview}, we simply add a softmax layer to the top of a contextual feature layer. Since our goal is not to make a good scene parsing network, our configuration is enough to regularize our main task.
Also, our objective function changes to the following
\begin{equation}
L = L_{reg} + \lambda L_{parse},
\end{equation}
where $L_{parse}$ is a cross-entropy loss of scene parsing task and $\lambda$ is a regularization weight.

\subsection{Implementation}
We implemented the proposed method using the TensorFlow running on a GeForce GTX 1080 GPU. With this setup, 500 epochs of training the network only takes several hours.

\paragraph{Data augmentation}
As the number of images in the dataset is small, the data augmentation is essential.
To generate more training data, we randomly rotate the input images from -10 to 10 degrees and flip horizontally.
We fill empty space by repeating pixel values of image boundaries to keep the dimension of image as 512$\times$512.
As mentioned, we adopt the sparse training method~\cite{BansalChen16,Larsson16} that randomly samples a few pixels for the backpropagation.
By doing this, we can generate many training examples from a small dataset.
In our implementation, we randomly choose 2048 pixels from an image for the sparse training.

\paragraph{Hyperparameters}
We train the proposed network using the Adam~\cite{kingma2014adam} optimization method with the learning rate of 1e-4 and the batch size of 4. The ResNet-50 layers are finetuned with 0.5x lower learning rate. We set $\alpha$ for training the semantic adjustment map to 0.8, $\delta$ of huber loss to 0.04, and $\lambda$ of cross-entropy loss of scene parsing task to 0.01 after the cross-validation. Determining the optimal number K is difficult as it is an unsupervised clustering problem. In our experiment, we found that 2, 4, and 2 for Foreground Pop-Out, Local Xpro, and Watercolor are sufficient for both the quantitative and qualitative result.

\section{Experiments}
\subsection{Dataset}
As mentioned, we use the dataset from~\cite{Yan16}, which is the only publicly available dataset for the semantics-aware photo adjustment. It contains 115 images from Flickr, of which the larger dimension is 512 pixels. In~\cite{Yan16}, the authors select 70 images for the training and the remaining 45 images for the testing. We use the same training and testing sets for a fair comparision. But, we additionally choose 10 images from the training set for the validation. Therefore, our training set is actually smaller than that of~\cite{Yan16}.

In the dataset, there are 3 types of photo adjustment effects: Foreground Pop-Out, Local Xpro, and Watercolor. For the Foreground Pop-Out effect, the contrast and the color saturation of foreground salient objects are increased while those of background objects are decreased. Local Xpro effect changes the brightness/contrast and the color of objects according to the predefined profiles for each semantic category. The adjustment of Watercolor is similar to that of Foreground Pop-Out except for an additional brush effect. In~\cite{Yan16}, the authors emulated the brush effect using superpixel segmentation~\cite{felzenszwalb2004efficient}. As our objective is to model spatially varing color mapping not texture, we follow the same procedure in~\cite{Yan16} for the brush effect.

\subsection{Baselines}
To show the effectiveness of the proposed method, we compare it with the method of Yan~\etal~\cite{Yan16}. As mentioned, we use the same training and the testing sets as described in~\cite{Yan16} except for the validation set. We also compare various design choices of the proposed method. For the easy reading, we name the proposed deep network as Semantics-Aware Adjustment Network (SA-AdjustNet), and we compare several variations of the SA-AdjustNet: SA-AdjustNet+MSE, SA-AdjustNet+Huber, SA-AdjustNet+Huber+MT, and SA-AdjustNet+Huber+MT+S. Each suffix after the name is the variation applied. MSE and Huber refer to the type of regression loss function, MT is the multi-task learning, and S indicates the network uses the semantic adjustment map as the contextual features. The networks without S use the convolutional features instead of the semantic adjustment map.

\begin{table}[t]
  \caption{Quantitative results. The values are $L_2$ distances in Lab color space.}
  \label{table:quantitative}
  \centering
  \begin{tabular}{llll}
    \toprule
                              & \multicolumn{3}{c}{Effects}                             \\
    \cmidrule{2-4}
                              & Foreground Pop-Out & Local Xpro       & Watercolor      \\
    \midrule
    Input                     & 13.86              & 19.71            & 15.30           \\
    \midrule
    Yan~\etal~\cite{Yan16}    &  7.08              &  7.43            &  7.20           \\
    \midrule
    SA-AdjustNet+MSE          &  7.16              &  7.06            &  6.92           \\
    SA-AdjustNet+Huber        &  6.59              &  6.97            &  6.81           \\
    SA-AdjustNet+Huber+MT     &  5.92              &  \textbf{6.66}   &  \textbf{6.75}  \\
    SA-AdjustNet+Huber+MT+S   &  \textbf{5.86}     &  7.03            &  6.83           \\
    \bottomrule
  \end{tabular}
\end{table}

\begin{figure}[h]
  \centering
  \hspace{-5.0mm}
  \subfigure[Input]
  {
    \begin{tabular}{l}
      \includegraphics[width=0.235\linewidth]{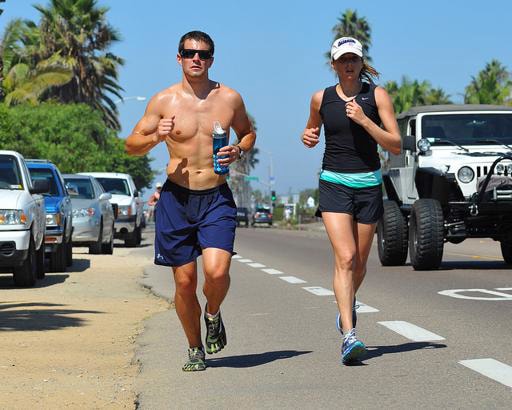} \\
      \includegraphics[width=0.235\linewidth]{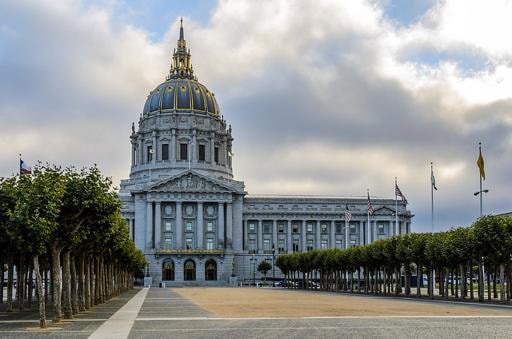} \\
      \includegraphics[width=0.235\linewidth]{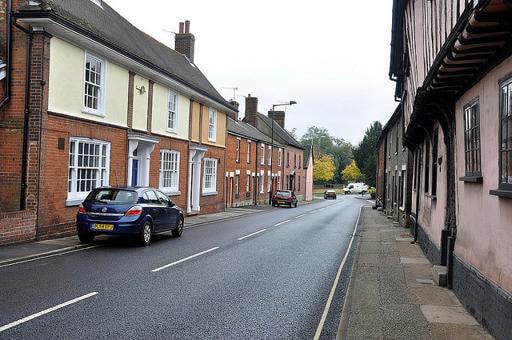}
    \end{tabular}
  }\hspace{-5.0mm}
  \subfigure[Yan~\etal~\cite{Yan16}]
  {
    \begin{tabular}{l}
      \includegraphics[width=0.235\linewidth]{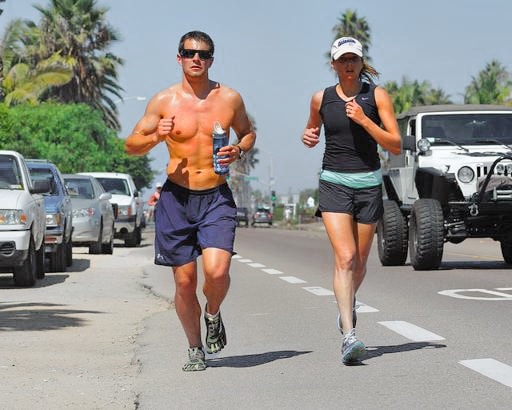} \\
      \includegraphics[width=0.235\linewidth]{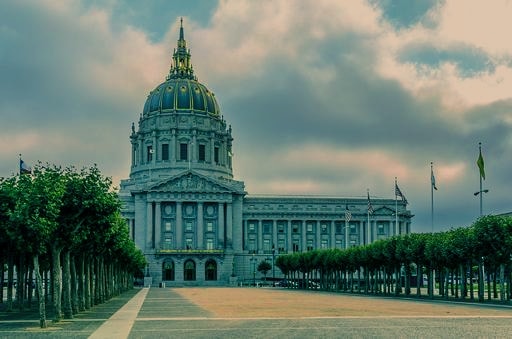} \\
      \includegraphics[width=0.235\linewidth]{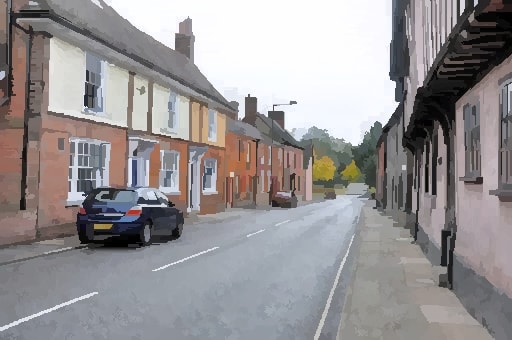}
    \end{tabular}
  }\hspace{-5.0mm}
  \subfigure[Ours]
  {
    \begin{tabular}{l}
      \includegraphics[width=0.235\linewidth]{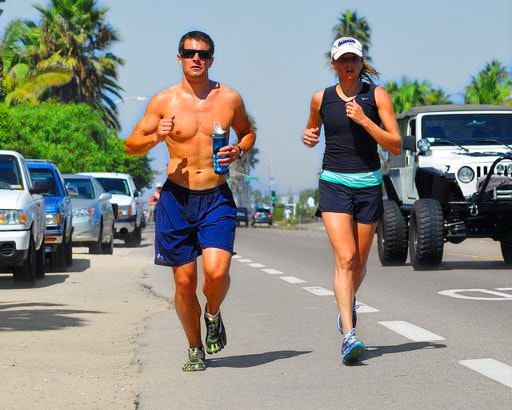} \\
      \includegraphics[width=0.235\linewidth]{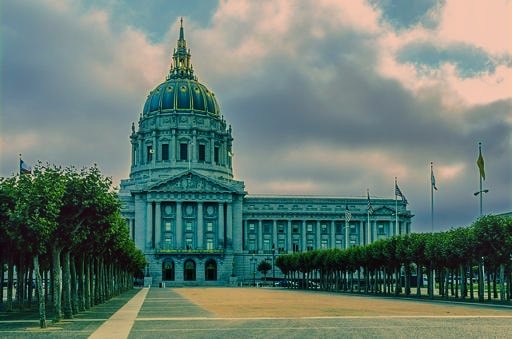} \\
      \includegraphics[width=0.235\linewidth]{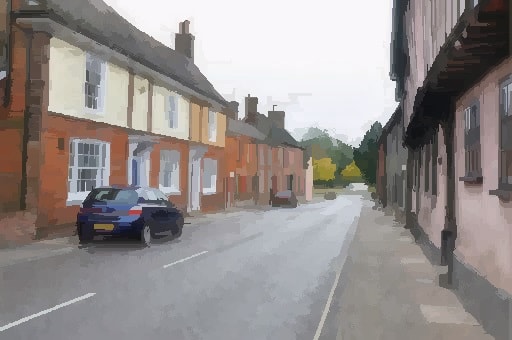}
    \end{tabular}
  }\hspace{-5.0mm}
  \subfigure[Ground truth]
  {
    \begin{tabular}{l}
      \includegraphics[width=0.235\linewidth]{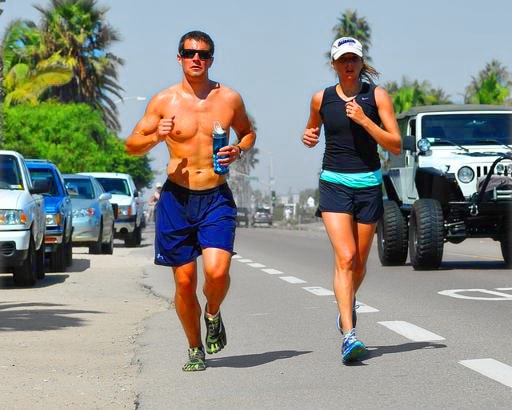} \\
      \includegraphics[width=0.235\linewidth]{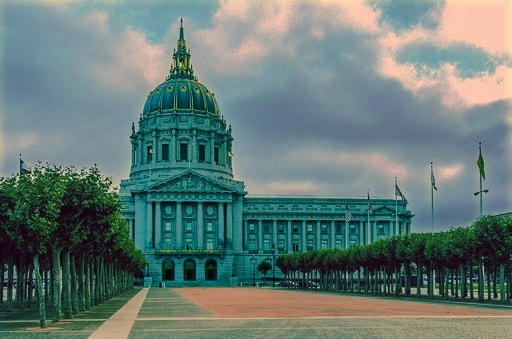} \\
      \includegraphics[width=0.235\linewidth]{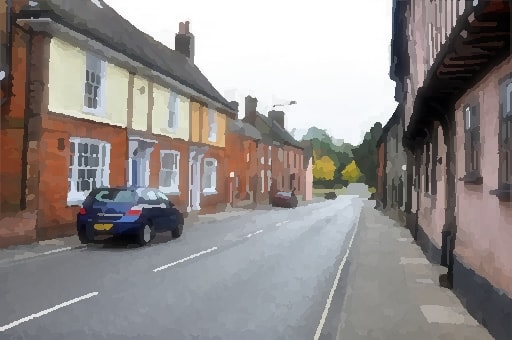}
    \end{tabular}
  }\hspace{-5.0mm}
  \caption{Qualitative results of SA-AdjustNet+Huber+MT. From top to bottom, the effects are Foreground Pop-Out, Local Xpro, and Watercolor, respectively.}
  \label{fig:qualitative}
\end{figure}

\begin{figure}[h]
  \centering
  \hspace{-5.0mm}
  \subfigure[Input]
  {
    \begin{tabular}{l}
      \includegraphics[width=0.235\linewidth]{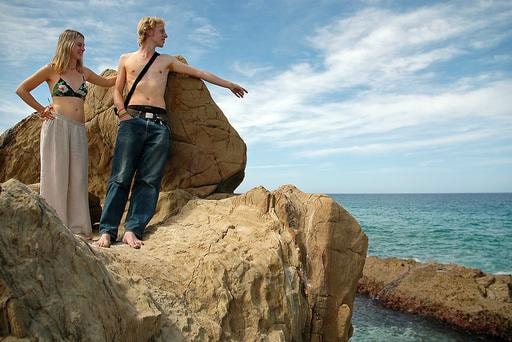} \\
      \includegraphics[width=0.235\linewidth]{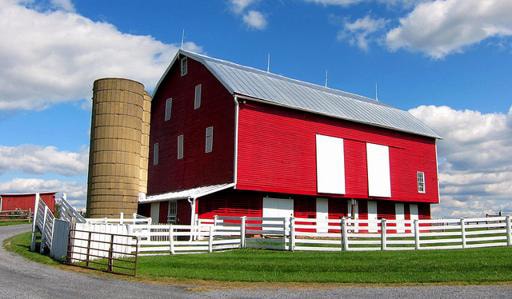} \\
      \includegraphics[width=0.235\linewidth]{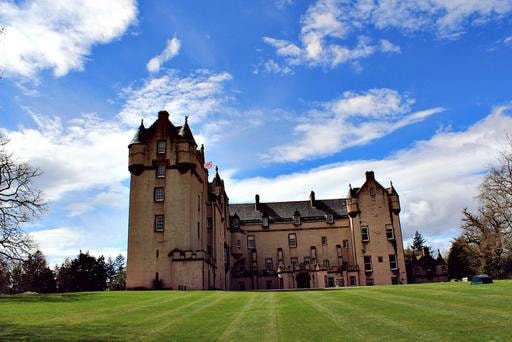}
    \end{tabular}
  }\hspace{-5.0mm}
  \subfigure[Ground truth]
  {
    \begin{tabular}{l}
      \includegraphics[width=0.235\linewidth]{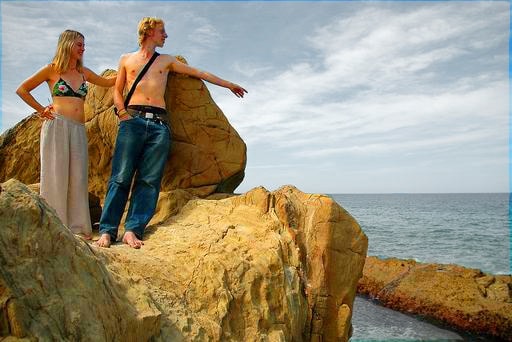} \\
      \includegraphics[width=0.235\linewidth]{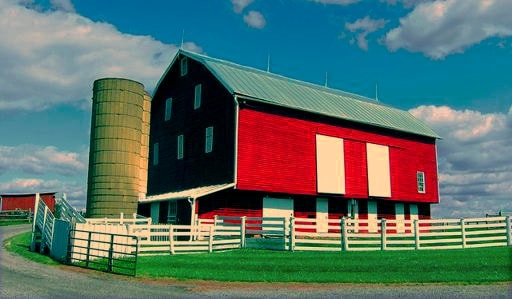} \\
      \includegraphics[width=0.235\linewidth]{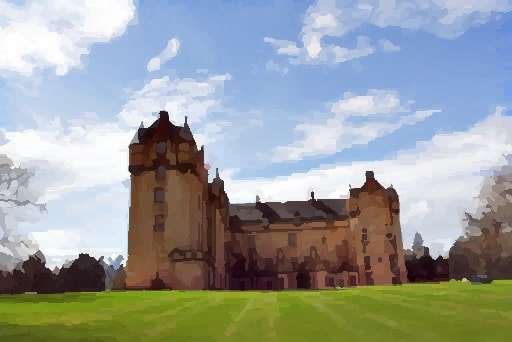}
    \end{tabular}
  }\hspace{-5.0mm}
  \subfigure[Ours]
  {
    \begin{tabular}{l}
      \includegraphics[width=0.235\linewidth]{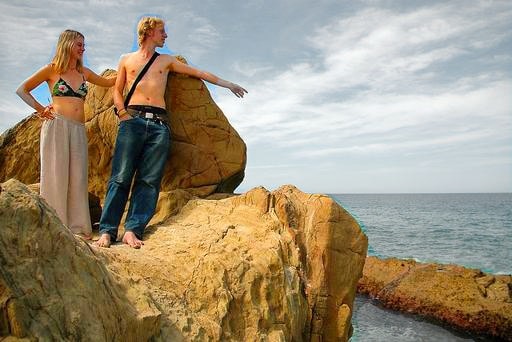} \\
      \includegraphics[width=0.235\linewidth]{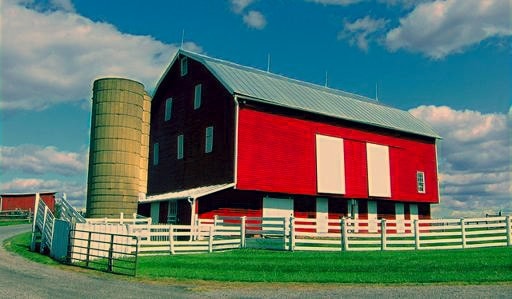} \\
      \includegraphics[width=0.235\linewidth]{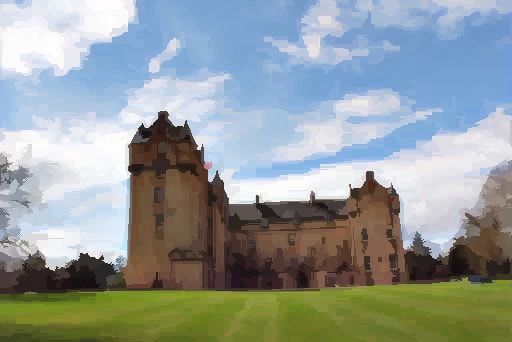}
    \end{tabular}
  }\hspace{-5.0mm}
  \subfigure[Semantic adjustment map]
  {
    \begin{tabular}{l}
      \includegraphics[width=0.235\linewidth]{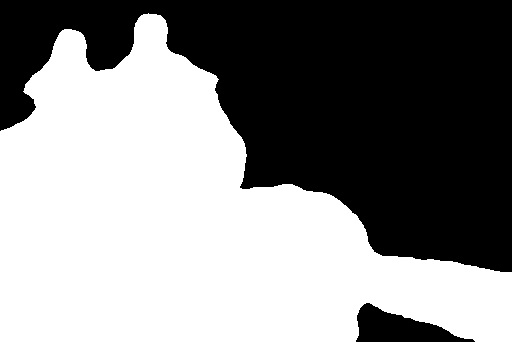} \\
      \includegraphics[width=0.235\linewidth]{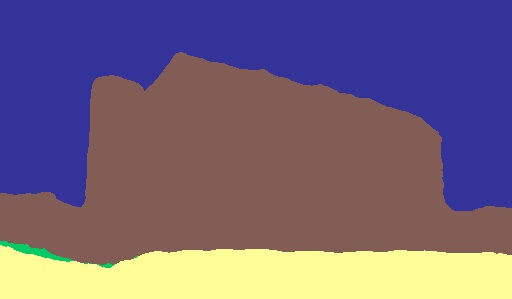} \\
      \includegraphics[width=0.235\linewidth]{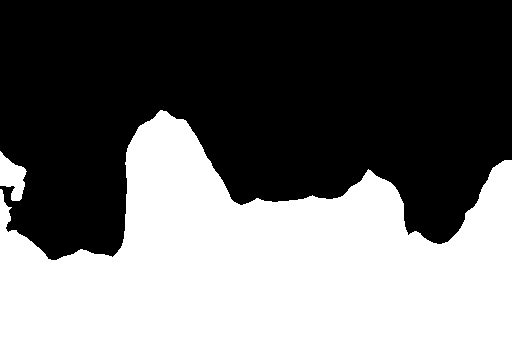}
    \end{tabular}
  }\hspace{-5.0mm}
  \caption{Some examples of the semantic adjustment map extracted from SA-AdjustNet+Huber+MT+S. The order of effects is same as~\fref{fig:qualitative}.}
  \label{fig:semantic_adjustment_map}
\end{figure}

\subsection{Experimental results}
\paragraph{Quantitative analysis}
\Tref{table:quantitative} shows the quantitative results of the proposed method. The values in the table are $L_2$ distance in the Lab color space. In most cases, the performance of the SA-AdjustNet is better than the method of~\cite{Yan16} since both the color and the contextual features of our method are jointly trained with the bilinear regression network. As shown in the table, the Huber loss and the multi-task learning are both effective for the regularization of the training of the proposed network. For the SA-AdjustNet+Huber+MT+S, the performance is competitive with that of the SA-AdjustNet+Huber+MT for the Foreground Pop-Out and Watercolor since the foreground and the background are balanced. However, the classes in the Local Xpro effect are diverse and imbalanced, and the optimal clustering is more difficult even if we use the class reweighting.

\begin{figure}[h]
  \centering
  \hspace{-5.0mm}
  \subfigure[Our estimation]
  {
    \begin{tabular}{l}
      \includegraphics[width=0.235\linewidth]{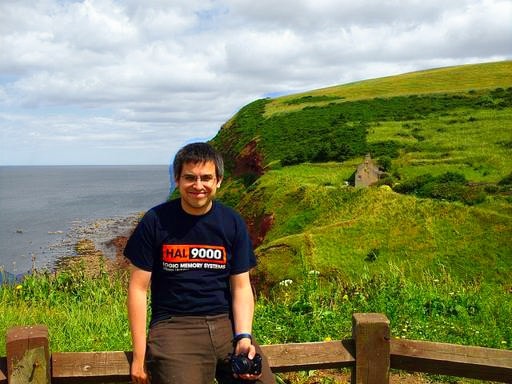} \\
      \includegraphics[width=0.235\linewidth]{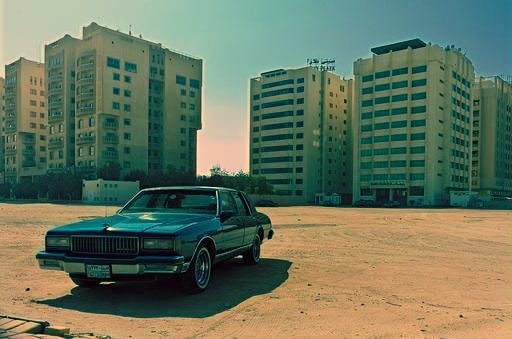}
    \end{tabular}
  }\hspace{-5.0mm}
  \subfigure[Semantic adjustment map]
  {
    \begin{tabular}{l}
      \includegraphics[width=0.235\linewidth]{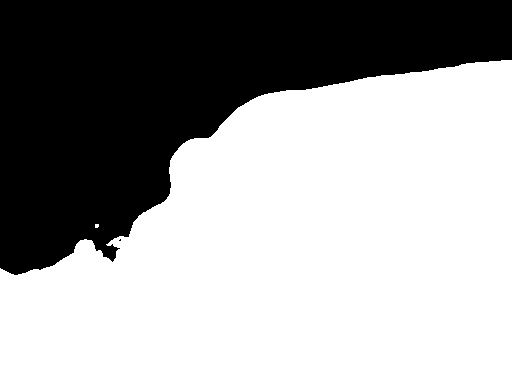} \\
      \includegraphics[width=0.235\linewidth]{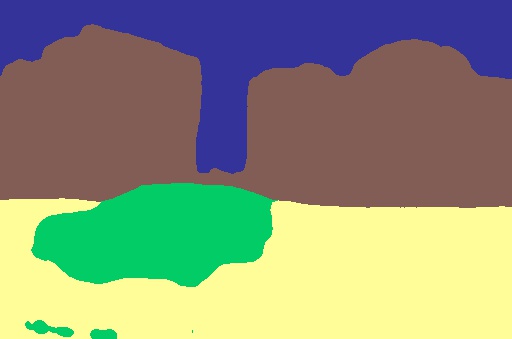}
    \end{tabular}
  }\hspace{-5.0mm}
  \subfigure[Personalized result]
  {
    \begin{tabular}{l}
      \includegraphics[width=0.235\linewidth]{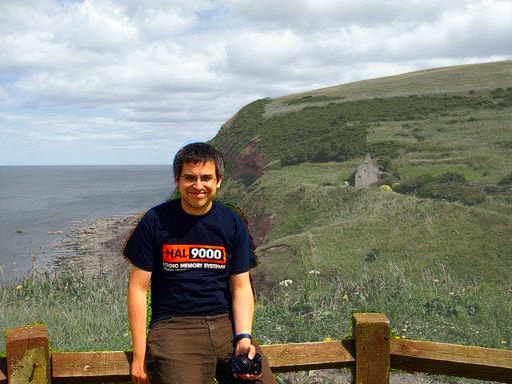} \\
      \includegraphics[width=0.235\linewidth]{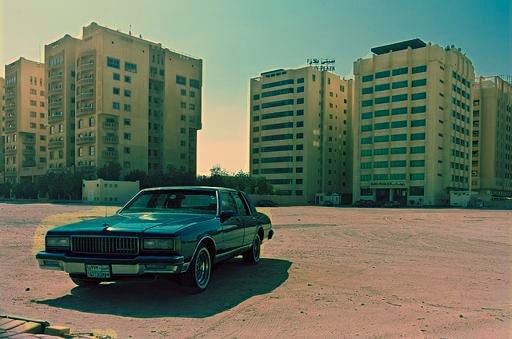}
    \end{tabular}
  }\hspace{-5.0mm}
  \subfigure[User preference map]
  {
    \begin{tabular}{l}
      \includegraphics[width=0.235\linewidth]{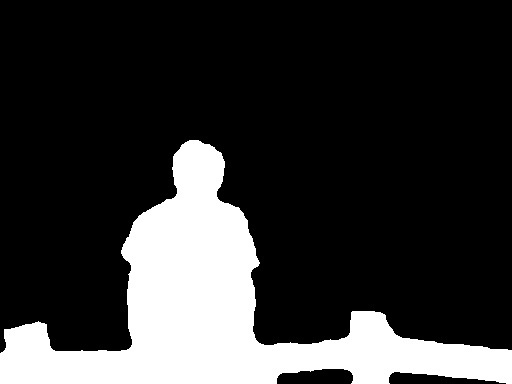} \\
      \includegraphics[width=0.235\linewidth]{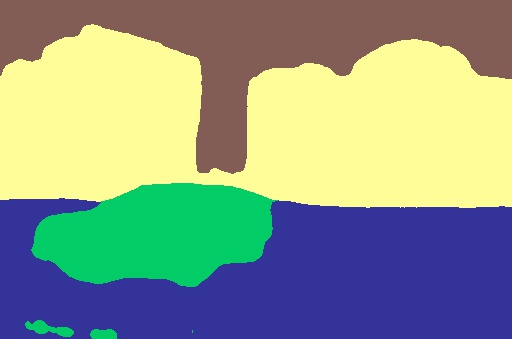}
    \end{tabular}
  }\hspace{-5.0mm}
  \caption{Some examples of personalized photo adjustment.}
  \label{fig:personalization}
\end{figure}

\paragraph{Qualitative analysis}
\Fref{fig:qualitative} shows some of the qualitative results from the test set. Each row of the figures show the 3 kinds of photo adjustment styles: Foreground Pop-Out, Local Xpro, and Watercolor. In most cases, the adjusted images using the proposed method are more visually pleasing and closer to the ground truth than those of Yan~\etal~\cite{Yan16}. As shown in the house of the 3rd row of~\fref{fig:qualitative}, the inconsistent color variation due to the incorrect segmentation is clearly reduced. \Fref{fig:semantic_adjustment_map} show some examples of the semantic adjustment map. The proposed network effectively discovers the inherent photo retouching styles. However, the semantic adjustment maps are discrete, and it results in the abrupt change of color around incorrect semantic boundaries as shown in the head of the man in~\fref{fig:semantic_adjustment_map}. This problem could be mitigated by considering neighborhood dependent models such as conditional random fields.

\subsection{Application: personalization of semantics-aware photo adjustment}
Although the proposed method provides the users with automatically adjusted photos, some users may want their photos to be retouched by their own preference. In the first row of~\fref{fig:qualitative} for example, a user may want only the color of the people to be changed. For such situations, we provide a way for the users to give their own adjustment maps to the system. \Fref{fig:personalization} shows some examples of the personalization. When the input image is forwarded, we substitue the extracted semantic adjustment map with the new adjustment map from the user. As shown in the figure, the proposed method effectively creates the personalized images adjusted by user's own style.

\section{Conclusion}
In this paper, we proposed a deep neural network for the semantics-aware photo adjustment. The proposed network learns the bilinear relationship between the color and the spatially varying scene context. With the semantic adjustment map, we can discover the inherent photo retouching presets within a style and apply it for the personalized photo adjustment. To effectively train the network, we use a robust loss function and the multi-task learning with the scene parsing task. The experimental results show that the proposed network outperforms an existing method both quantitatively and qualitatively.

% This work was supported by Global Ph.D. Fellowship Program through the National Research Foundation of Korea (NRF) funded by the Ministry of Education (NRF-2015H1A2A1033924).

{\small
\bibliographystyle{ieeetr}
\bibliography{egbib}
}

\end{document}